\documentclass[conference]{IEEEtran}
\IEEEoverridecommandlockouts
\usepackage{cite}
\usepackage{amsmath,amssymb,amsfonts}
\usepackage{graphicx}
\usepackage{textcomp}
\usepackage{xcolor}
\usepackage{hyperref}
\usepackage{multicol}
\usepackage{algorithm,algpseudocode}
\usepackage{geometry}

\def\BibTeX{{\rm B\kern-.05em{\sc i\kern-.025em b}\kern-.08em
    T\kern-.1667em\lower.7ex\hbox{E}\kern-.125emX}}
    \makeatletter
\newcommand{\linebreakand}{%
  \end{@IEEEauthorhalign}
  \hfill\mbox{}\par
  \mbox{}\hfill\begin{@IEEEauthorhalign}
}

\usepackage{afterpage}
\afterpage{
  \ifnum\value{page}=4 
    \enlargethispage{-2.02in} 
  \fi
}

\begin{document}

\title{Large Language Models for Video Surveillance Applications}

    

\author{
    \IEEEauthorblockN{Ulindu De Silva\textsuperscript{1}}
    \IEEEauthorblockA{
        \textit{Dept. of ENTC} \\  
        \textit{University of Moratuwa}\\
        Sri Lanka\\
        \href{mailto:desilvaalup.20@uom.lk}{desilvaalup.20@uom.lk}
    }
    \and
    \IEEEauthorblockN{Leon Fernando\textsuperscript{1}}
    \IEEEauthorblockA{
        \textit{Dept. of ENTC} \\
        \textit{University of Moratuwa}\\
        Sri Lanka\\
        \href{mailto:fernandoknal.20@uom.lk}{fernandoknal.20@uom.lk}
    }
    \and
    \IEEEauthorblockN{Billy Lau Pik Lik}
    \IEEEauthorblockA{
        \textit{Engineering and Product Development} \\
        \textit{SUTD} \\
        Singapore\\
        \href{mailto:billy_lau@sutd.edu.sg}{billy\_lau@sutd.edu.sg}
    }
    \linebreakand
    \IEEEauthorblockN{Zann Koh}
    \IEEEauthorblockA{
        \textit{Engineering and Product Development}\\
        \textit{SUTD}\\
        Singapore\\
        \href{mailto:zann_koh@sutd.edu.sg}{zann\_koh@sutd.edu.sg}
    }
    \and 
    \IEEEauthorblockN{Sam Conrad Joyce}
    \IEEEauthorblockA{
        \textit{Architecture and Sustainable Design}\\
        \textit{SUTD}\\
        Singapore\\
        \href{mailto:sam_joyce@sutd.edu.sg}{sam\_joyce@sutd.edu.sg}
    }
    \and 
    \IEEEauthorblockN{Belinda Yuen}
    \IEEEauthorblockA{
        \textit{Lee Kuan Yew Centre for Innovative Cities}\\
        \textit{SUTD}\\
        Singapore\\
        \href{mailto:belinda_yuen@sutd.edu.sg}{belinda\_yuen@sutd.edu.sg}
    }
    \linebreakand
    \IEEEauthorblockN{Chau Yuen}
    \IEEEauthorblockA{
        \textit{School of Electrical and Electronic Engineering} \\
        \textit{Nanyang Technological University} \\
        Singapore\\
        \href{mailto:chau.yuen@ntu.edu.sg}{chau.yuen@ntu.edu.sg}
    }
}

\maketitle

\footnotetext[1]{Equal Contribution. Work done while interning at SUTD, Singapore.}

\begin{abstract}
The rapid increase in video content production has resulted in enormous data volumes, creating significant challenges for efficient analysis and resource management. To address this, robust video analysis tools are essential. This paper presents an innovative proof of concept using Generative Artificial Intelligence (GenAI) in the form of Vision Language Models to enhance the downstream video analysis process. Our tool generates customized textual summaries based on user-defined queries, providing focused insights within extensive video datasets. Unlike traditional methods that offer generic summaries or limited action recognition, our approach utilizes Vision Language Models to extract relevant information, improving analysis precision and efficiency. The proposed method produces textual summaries from extensive CCTV footage, which can then be stored for an indefinite time in a very small storage space compared to videos, allowing users to quickly navigate and verify significant events without exhaustive manual review. Qualitative evaluations result in 80\% and 70\% accuracy in temporal and spatial quality and consistency of the pipeline respectively.
\end{abstract}


\section{Introduction}
In recent years, unprecedented advancements in hardware have been propelled by companies like NVIDIA. These advances have significantly increased processing power, enabling the analysis of large amounts of data generated by the Internet. This era of big data allows for the discovery of patterns that were previously undetectable. The high-tech companies have invested billions in research, resulting in the development of highly sophisticated models. Among these are large-scale vision-language models\cite{hong2024cogagent}\cite{bai2023qwen}\cite{chen2022pali}\cite{zhu2023minigpt}. Flamingo\cite{alayrac2022flamingo} proposes gated cross-attention layers which enable the model to accept arbitrarily interleaved visual data and text as input to generate text in an open-ended manner. BLIP-2\cite{li2023blip} proposes a Q-Former trained in two stages to bridge the modality gap which enables zero-shot instructed image-to text generation. InternVL\cite{chen2024internvl} is the latest model with state-of-the-art results in visual perception tasks. 

From smartphones to CCTV cameras, numerous devices now possess the capability to record video. While videos carry high semantic value, they are costly to store and analyze. Storing more requires more hardware and thus more financial resources. Therefore, extracting video data into textual format is crucial for efficient analysis, sharing, and storage.

\begin{figure*}[h]
    \centering
    \centerline{\includegraphics[width=1\linewidth]{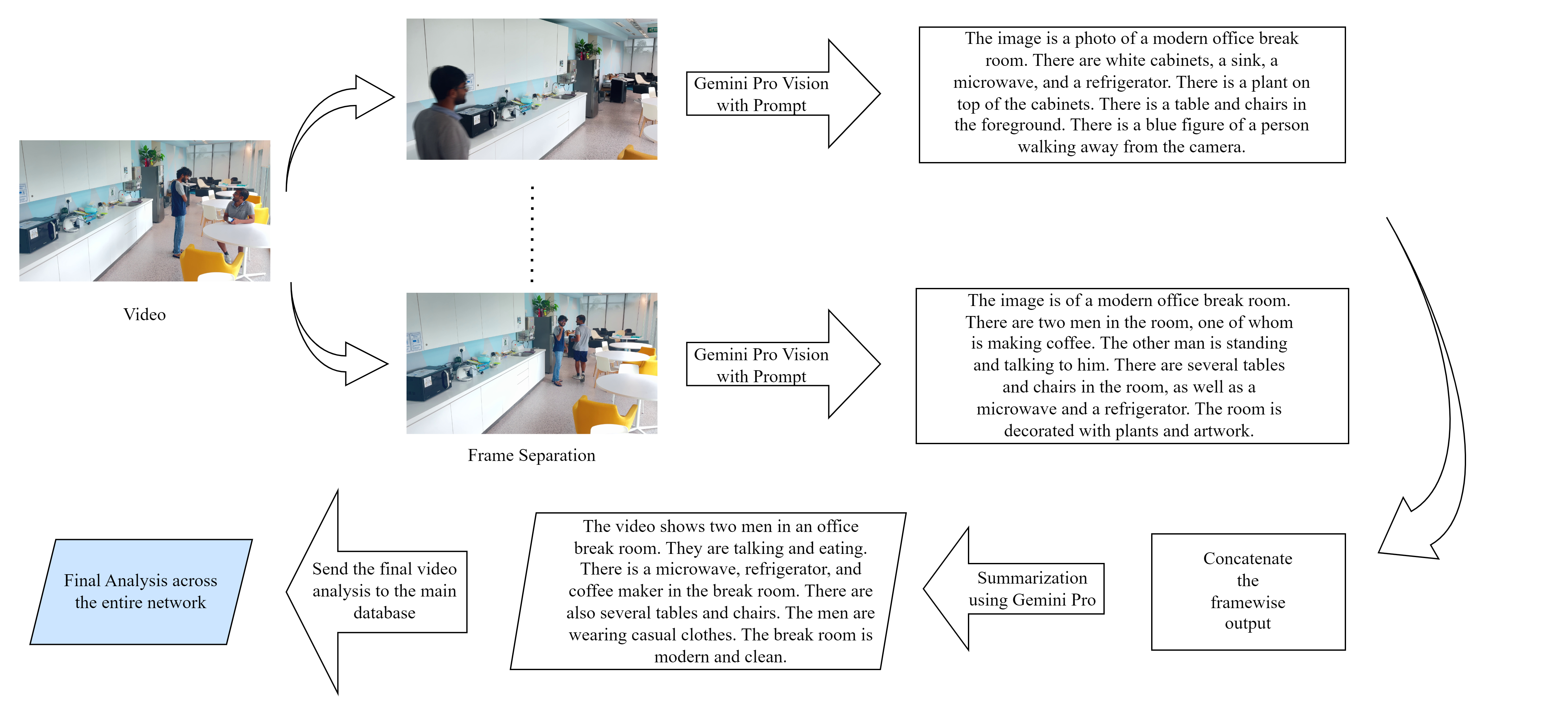}}
    \caption{Proposed Method Overview}
    \label{fig:1}
\end{figure*}

The advent of GenAI has transformed numerous industries by presenting unparalleled optimization opportunities. GenAI \cite{radford2018improving}\cite{fui2023generative} technologies, underpinned by advanced deep learning models, have exhibited extraordinary proficiency in diverse applications, such as natural language processing \cite{iorliam2024comparative} and image generation \cite{chang2023muse}\cite{rombach2022high}. The capabilities of these GenAI-based tools to handle vast volumes of real-time data and derive actionable insights have established them as essential assets for businesses aiming to maintain a competitive edge in the modern marketplace.

Our study focuses on CCTV networks with multiple cameras integrated into a unified system. Traditionally, each camera captures data independently and sends it to a central hub for manual anomaly inspection. This paper proposes a novel video analysis pipeline using vision language models. Unlike conventional methods, which provide generic summaries or recognize limited actions, our approach offers tailored textual summaries based on user-defined queries. By filtering out irrelevant data, our method enhances video exploration and retrieval efficiency, meeting specific user needs. We analyze CCTV footage both individually and collectively, providing a comprehensive surveillance perspective. The primary contributions of this paper include;
\begin{enumerate}
    \item Introduction of an innovative pipeline for user-friendly automated video analysis in a network.
    \item Implementation of our proposed methodology on a benchmark dataset and real-world videos.
    \item Analysis of the effectiveness of our method in temporal and spatial understanding of videos.
    \item Comprehensive qualitative evaluation to validate the methodology's effectiveness and accuracy.
\end{enumerate}

The rest of the paper is organized as follows: Section \textrm{II} discusses the proposed methodology followed by Section \textrm{III} with the results and discussion. The paper ends with the Section \textrm{IV} conclusion and future works. 

\section{Methodology}
In this section, we introduce our innovative pipeline for video analysis within a video network. The initial phase of the analysis is conducted independently for each video camera. As in Figure [\ref{fig:1}], the video stream is initially segmented into individual frames based on a predetermined frame rate, which is customized for the specific use case. This adjustable frame rate provides flexibility, allowing for a higher frame rate for fast-moving content and a lower frame rate for slower-paced content.

\begin{figure}[h]
    \centering
    \centerline{\includegraphics[width=1\linewidth]{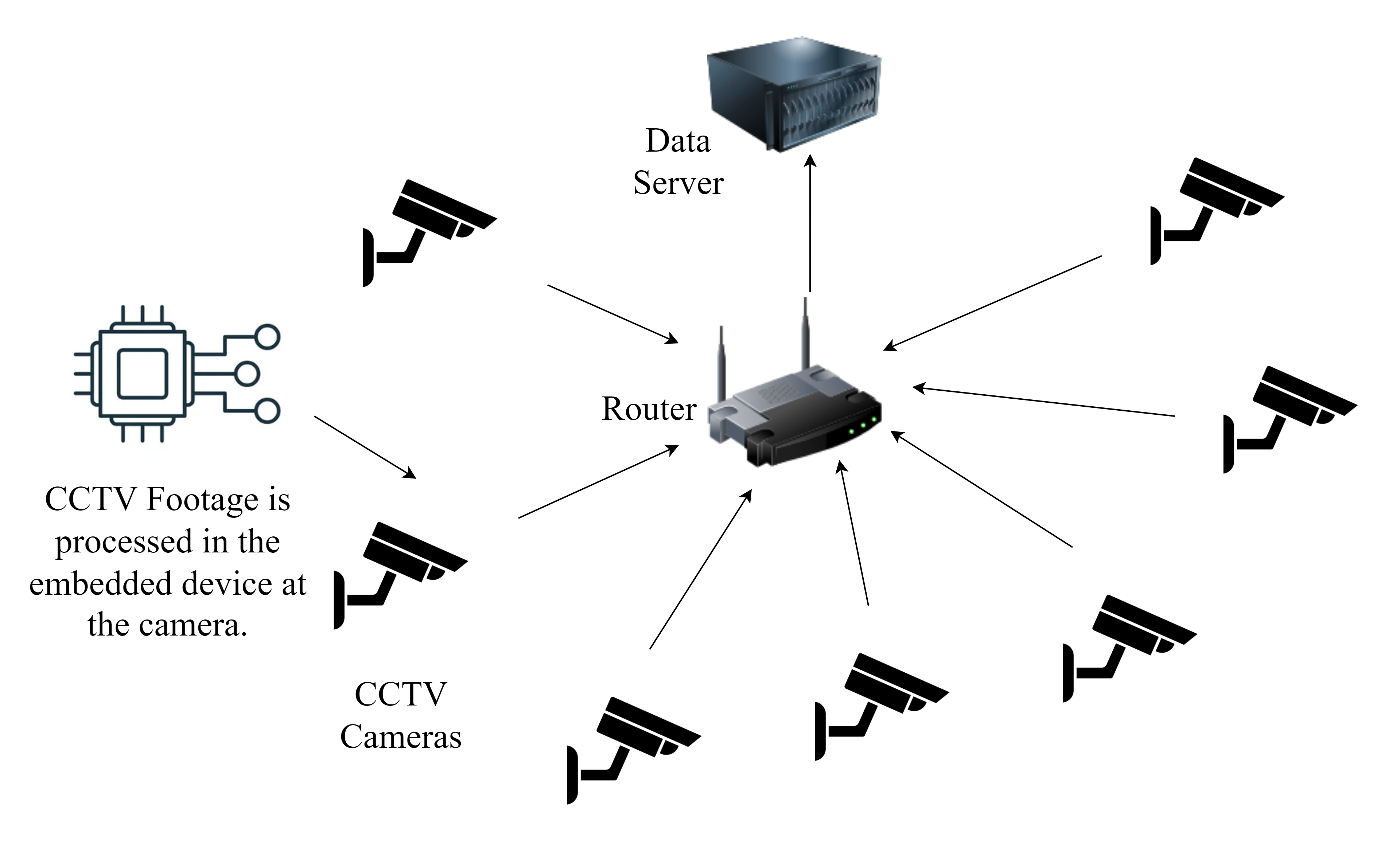}}
    \caption{CCTV Network}
    \label{fig:2}
\end{figure}

The segmented frames are then processed using the GenAI model, specifically Gemini 1.0 Pro Vision. \cite{team2023gemini} The model's ability to respond to contextual nuances is enhanced by adjusting the temperature variable, providing flexibility in generating content that meets specific requirements. In addition, the prompt can be customized to align with the desired output. Based on the expected results, we adjust the safety settings to address potential issues related to harm categories, including harassment, hate speech, sexual content, and dangerous content. In our prompts, we instruct Gemini Pro Vision to describe each image, and this process is repeated for all frames in the video, ensuring comprehensive and detailed analysis.

However, the frame-by-frame output tends to be noisy and redundant, as minimal changes typically occur between consecutive frames. Consequently, many of the descriptions for adjacent frames are similar, making it inefficient to store each one. This highlights the necessity for a filtering mechanism. To address this, we leverage the capabilities of Gemini Pro. By concatenating the outputs from all frames and feeding them into the model, we can efficiently filter and streamline the data, eliminating redundancies and preserving only the most relevant information. The final output provides a comprehensive summary of the entire video, directing the user to key sections that require attention. This filtering mechanism is achieved through a carefully crafted prompt given to Gemini Pro during the summarization task. This video analysis is performed on the CCTV camera and transferred to the main database for further processing and fusion with the output achieved from other CCTV cameras on the network, as shown in Figure [\ref{fig:2}].


The summarized video descriptions from all the CCTVs in the network are stored in the database. Next, we input the concatenated video summaries into Gemini Pro to generate an overarching summary throughout the network. This comprehensive summary provides a holistic understanding of events within the vicinity of the CCTV network during the specified period. By synthesizing information from multiple sources, we can identify patterns, correlate events, and obtain a complete picture of the monitored area, enhancing situational awareness and decision-making.

The standard pipeline can be modified to accommodate specific use cases. For instance, to analyze the behaviour of a particular individual, we can query the database to identify if such a person is described in the video summaries. These descriptions can then be manually reviewed to gain insights. Alternatively, the system can be queried to generate a summary that tracks the individual through both space and time, providing a comprehensive behavioural analysis. This adaptability of the pipeline allows for precise, user-defined inquiries, making it a versatile tool for various applications such as security, behavioural studies, and customer analysis. Using the power of generative AI, the system ensures that the information extracted is both relevant and actionable, significantly enhancing the efficiency and precision of the analysis.

\section{Experiments and Results}

\subsection{Video Dataset}
Microsoft Research-Video to Text, MSR-VTT  \cite{xu2016msr} is a large-scale video benchmark dataset which contains more than 10K web video clips and 200K clip-sentence pairs consisting of 20 categories. We used selected videos from the dataset to experiment with the proposed model and evaluate its performance.

Additionally, we curated a custom dataset by selecting YouTube CCTV videos from supermarket environments and roadside footage capturing accidents. Furthermore, we included videos of individuals moving, talking, and interacting within the Singapore University of Technology and Design (SUTD) campus. These videos encompass a wide array of scenarios involving human behaviour in retail settings and various traffic incidents, providing a diverse and comprehensive dataset for our analysis.

\begin{figure}[h]
    \centering
    \centerline{\includegraphics[width=1\linewidth]{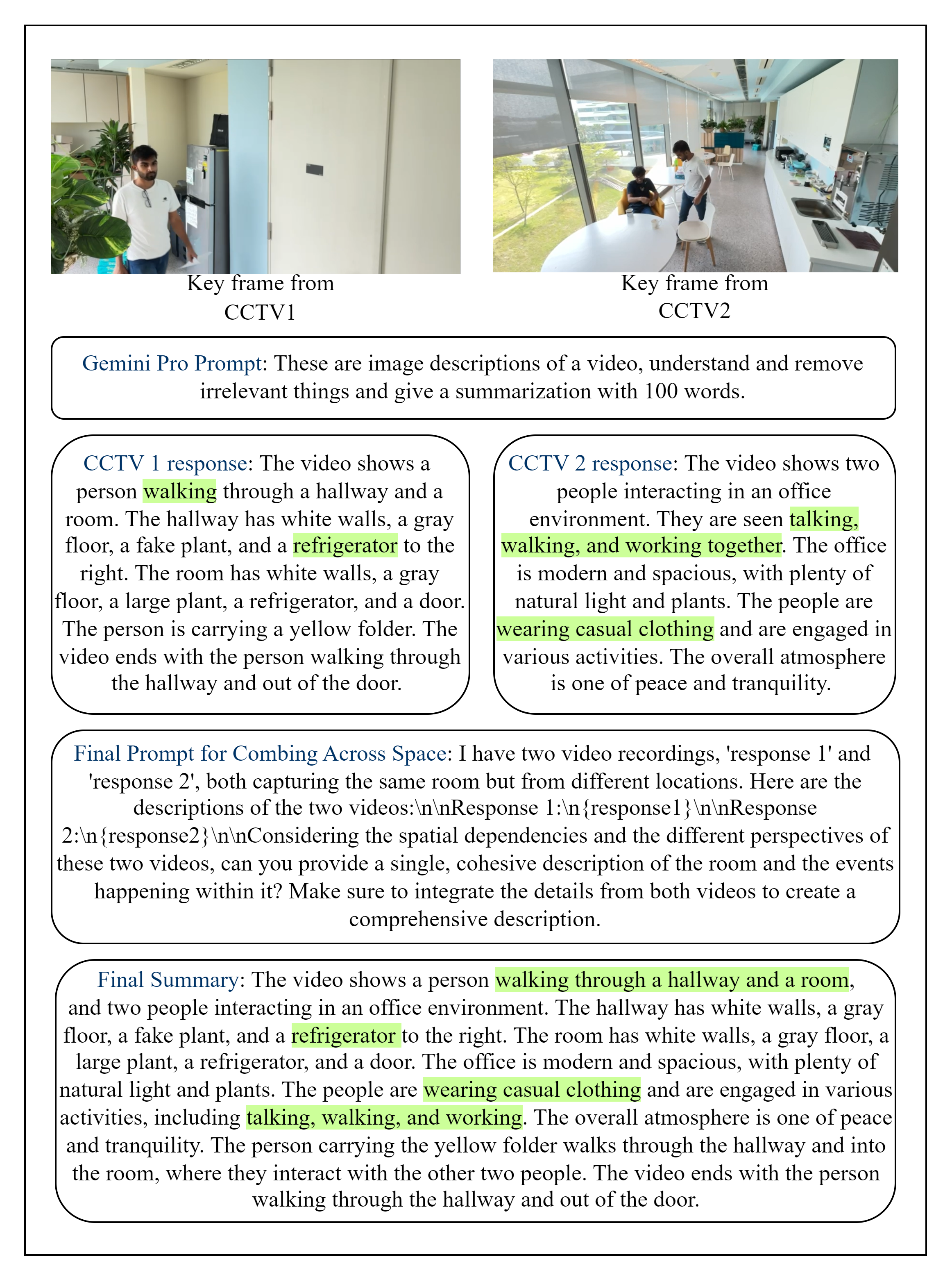}}
    \caption{Temporal and Spatial analysis for two sample videos taken in a room from two different viewpoints one after the other. Highlighted in light green are occurrences when our method was able to detect information such as people, interactions, and environmental details which helps in creating a vivid picture of the scene.}
    \label{fig:3}
\end{figure}

\subsection{Qualitative Analysis}
In this section, we present a qualitative analysis of the image description capabilities of the Gemini Pro Vision model. Our evaluation demonstrates that the model can effectively describe images, capturing the most important elements without explicit guidance. As illustrated in Figure [\ref{fig:3}], the model successfully identifies key aspects such as people, their interactions, and relevant environmental details, while appropriately omitting less significant elements like the number of chairs and tables in the background.

Despite these strengths, the model occasionally makes errors, particularly in colour identification. Additionally, even with temperature values set close to zero to minimize creativity, the model sometimes generates descriptions of elements that are not present in the image.

To conduct a comprehensive temporal and spatial qualitative analysis, we enlisted the assistance of 10 students from the Faculty of Engineering, University of Moratuwa. These students were tasked with evaluating the generated summaries and scoring the model's performance on a scale from 1 to 10. This evaluation focused on the model's ability to accurately and effectively describe the images in various contexts, providing a more in-depth understanding of its strengths and limitations. 

The qualitative analysis includes the extensive evaluation of temporal and spatial consistency of the generated output, which will be presented in detail in the next two subsections.\\

\textbf{Temporal Consistency.}
When analyzing a video by splitting it into individual frames, one can observe various actions and changes occurring over time. The goal of evaluating temporal consistency is to determine if our framework can accurately capture and represent these temporal events.

The Gemini Pro model excels in this regard, successfully identifying the chronological order of frames and recognizing when no significant action occurs. For instance, if the same description is generated for consecutive frames, it indicates that no notable action has taken place. Conversely, when a person is seen standing in one frame and then sitting in another, Gemini Pro accurately outputs that the person has moved and sat down.

Our evaluators, as mentioned in the previous section, rated the temporal consistency of Gemini Pro's summaries an average of 8 out of 10. This high rating reflects the model's strong ability to capture and represent the sequence of events accurately. \\

\textbf{Spatial Consistency.}
A single camera's field of view is inherently limited, making it insufficient for applications requiring comprehensive coverage. In such cases, analyzing footage from individual cameras may not provide a complete understanding of the overall situation, thus highlighting the need for video network analysis.

In this section, we evaluate how our framework maintains spatial consistency across multiple videos. We conducted tests using five pairs of videos captured from two different positions, each offering distinct points of view. The model demonstrated its capability to effectively merge the perspectives, accurately representing the spatial relationships across different areas.

The evaluators provided an average score of 7 out of 10 for spatial consistency. This indicates that while the model performs well in combining footage from multiple cameras, there is room for improvement in certain scenarios.

\section{Conclusion and Future Works}
This paper introduces a proof of concept novel pipeline for video analysis aimed at reducing the time and cost associated with traditional methods, where entire videos must be manually evaluated. Unlike anomaly detection or action recognition models that rely on discrete sets of actions, our approach dynamically generates comprehensive descriptions. Experimental results confirm the model's capability to accurately depict video content. Qualitative analysis also demonstrates the framework's proficiency in temporal and spatial video reconstruction using customizable prompts.

Relying on cloud-based models is not commercially viable due to the costs of API calls and the latency involved in uploading images and retrieving outputs. A practical alternative is to deploy the model on an onboard computer or a local server, especially if the model is too complex for a local machine. Future work includes creating a comprehensive surveillance dataset with ground truth annotations to rigorously evaluate the model's performance and fine-tune the model with domain-specific images.

\bibliographystyle{ieeetr}
\bibliography{main}





\end{document}